# Live-Wire 3D Medical Images Segmentation

## Ognjen Arandjelovic

3$^{rd}$ year ECS project, Engineering Department, Oxford University, UK

# Live-Wire 3D Medical Images Segmentation

## Ognjen Arandjelovic[1]

3rd year ECS project, Engineering Department, Oxford University, UK

## Abstract

This report describes the design, implementation, evaluation and original enhancements to the Live-Wire method for 2D and 3D image segmentation. Live-Wire 2D employs a semi-automatic paradigm; the user is asked to select a few boundary points of the object to segment, to steer the process in the right direction, while the result is displayed in real time. In our implementation segmentation is extended to three dimensions by performing this process on a slice-by-slice basis. User's time and involvement is further reduced by allowing him to specify object contours in planes orthogonal to the slices. If these planes are chosen strategically, Live-Wire 3D can perform 2D segmentation in the plane of each slice automatically. This report also proposes two improvements to the original method, path heating and a new graph edge feature function based on variance of path properties along the boundary. We show that these improvements lead up to a 33% reduction in interaction with the user, and improved delineation in presence of strong interfering edges.

## 1. INTRODUCTION

Segmentation of medical images refers to the process of identifying regions of similar tissues in images. This is usually the first step when further quantitative analysis of anatomical structures is needed. Applications making use of it range from simple ones like volume estimation, to more complex ones, such as morphological abnormality recognition. This wide and frequent demand for it, emphasises the need for fast and accurate methods of 3D segmentation.

However, inherent underlying problems contained in this task, despite continuous efforts put into solving it, still mean that fully automatic segmentation has not been solved. Imaging methods, like magnetic resonance imaging (MRI) or ultrasound scans, typically produce low contrast or noisy images, with sometimes anisotropic distortions [11]. In addition, the properties of the scanned tissues result in their variable density and shape, making a priori information of limited use. In connection with this, the performance of the methods developed critically depends on the features to be detected, making them very sensitive to variations in input data.

[1] E-mail address: Ognjen.Arandjelovic@sjc.ox.ac.uk

In the literature, the most commonly used method of medical image segmentation has been manual segmentation[10]. This method typically requires the user to segment the object of interest by drawing its contour manually. Due to its high dependence on human effort, this method has become unacceptable with increase of computer power. This is especially evident in 3D segmentation, which typically involves 20-40 plane objects to be segmented. On top of that, manual segmentation tends to produce inaccurate and inconsistent results making it a choice of preference in very few cases.

Recognition of difficulties associated with fully automatic and manual segmentation has resulted in a number of semi-automatic (also known as interactive) methods being developed over the course of the last decade[10]. These rely on user-computer interaction as a means of obtaining additional information about the data, allowing for increased accuracy. The choice of the information to be specified by the user is dictated by the paradigm employed by the method in question and is obviously a step of crucial importance. This is explored further in section 2.2. In particular, the subject of my project is the Live-wire 3D method for 3D image segmentation, that relies on a number of user selected object boundary points, for steering the process in the right direction. Through the analysis of the original algorithm we identify the major limitation of it and investigate a number of ways of improving accuracy, repeatability and reduction in user effort required for segmentation.

This report is organised as follows. The basic Live-Wire 2D method and its extensions are described in section 2. Section 3 explains one way of extending this method to three dimensions. Section 4 deals with design and implementation issues. Experimental results are presented in section 5 and conclusions are made in section 6.

## 2 LIVE-WIRE 2D

Live-Wire 2D boundary detection is a general method for user-assisted segmentation of images[1]. Its use is not restricted to medical imaging (indeed, it was first developed for the purpose of image manipulation; see [1]). This project focuses on its application to 3D medical image segmentation, with evaluation on MRI data, such as shown in Figure 1.

In this section I present the Live-Wire method for planar (2D) segmentation. First the basic method is described, followed by some ideas on its improvement. In section 3, I will present one new way of extending this method to three dimensions I have developed and outline some key difficulties in this step.

## 2.1 Basic live-wire 2D method

In two dimensions, from a user's point of view, live-wire 2D consists of selecting a *seed point*, a point on the boundary of the object that we want to segment, and moving the mouse pointer to another location on the boundary. The estimate of the boundary segment from the seed point to the current location of the cursor, a *live-wire*, is displayed in real time. Once satisfactory estimate is displayed, the user clicks the mouse and this segment estimate becomes a part of the global boundary estimate. The end point of the segment is subsequently used as a new seed point and the process is repeated until the boundary is closed. This process is shown in figure 1.

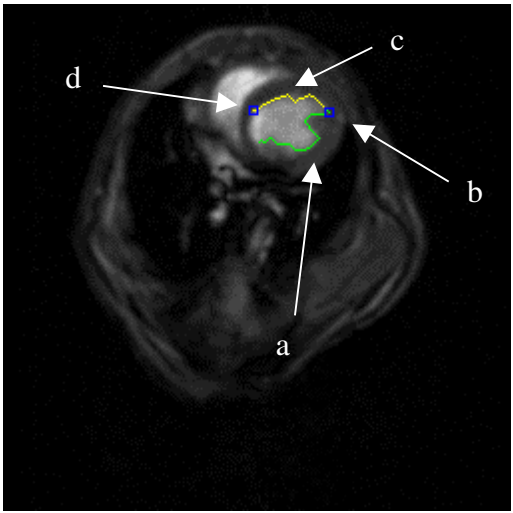

Figure 1.

Live-Wire 2D segmentation of a mouse ventricle. (a) segmented part of the boundary, (b) current seed point, (c) live-wire, (d) position of the cursor.

In the Live-Wire algorithm, the estimate of an object's boundary is achieved by considering the image as an undirected, connected graph with pixels as nodes and edges connecting neighbouring pixels (8 neighbourhood). The edges to a pixel are assigned weights depending on the estimated likelihood of them being on an object edge. Consequently, a boundary estimate span between two points can be calculated using Dijkstra's algorithm[3, 13] where the shortest path corresponds to the most likely boundary.

Obviously, the method of assigning weights to the edges is crucial for the performance of Live-Wire. The goal is to identify features of the edge pixels and use them in estimating the probability of a pixel belonging to an edge. Two feature weighing functions initially proposed in the original work are gradient magnitude and Laplacian zero-crossing. The gradient feature function scales the intensity gradient to the working intensity range (typically 0-255) and inverts it so that high gradients produce low cost edges. Laplacian zero-crossing based feature function produces low cost (close to 0) for pixels that correspond to a vanishing second derivative, and high (close to 255) for others. The weight assigned to the pixel edge is then calculated using a linear combination of these two.

$$f_G = 255*(1 - \frac{|\nabla I|}{grad_{MAX}}) \qquad f_L = \begin{cases} 1, \ \nabla^2 I = 0 \\ 255, \ \nabla^2 I \neq 0 \end{cases}$$

$$edge = w_G * f_G + w_L * f_L \qquad \text{Where } I \text{ is the image intensity.}$$

Based on empirical evidence, Barrett et al.[1], suggest equal weighting of the two feature functions. In section 2.3, I will present several modifications to this idea, some original, that result in further improvement of the results.

## 2.2 Key Live-Wire 2D aspects

The basic method described in section 2.1 shows several important aspects of Live-Wire in which it differs from other semi-automatic segmentation methods in frequent use, such as the Snakes method.

First of all, Live-Wire 2D (through this report we will occasionally refer to it as Live-Wire instead of Live-Wire 2D) method works in a piece-wise manner. This means that the boundary created with Live-Wire is not globally optimal, but piece-wise optimal. This important distinction shows to be beneficial in practice. Methods that optimise a global fitness function of the boundary, usually by minimising a certain energy function, tend to produce artificial oscillations[1]. Live-Wire does not have this weakness.

Live-Wire also brings innovations in the way user-software interaction is performed. It is easy and intuitive to use, and importantly, the result is displayed to the user in real-time before it is accepted as final, allowing for further adjustments. This manner of interaction also has the benefit of allowing the user to adjust his actions to the method performance on particular data, for example by selecting shorter segments in the case of noisy or low contrast images. In contrast, methods that produce a globally optimal boundary do not have this property. The user is unaware of the shape of the result until it is computed, and in the case of an unsatisfactory result the whole procedure has to be repeated. Even more importantly, the way the input should be modified after unsatisfactory results have been produced is unclear and unintuitive.

These observations suggest that Live-Wire can turn out to be a powerful and useful method. Section 6 will explore these questions in more depth and on MRI data.

## 2.3 Improvements to Live-Wire 2D

After the original method had been implemented, and shown to perform well in practice (see section 5), a number of improvements to it have been investigated. Among these, it is instructive to differentiate between two classes. One class comprises improvements that are intrinsic to the underlying computational method, while the improvements of the second monitor for certain patterns in user behaviour and try to use this information to predict user's intentions. Both are examined further in the following sections.

### 2.3.1 Main challenges

For the purpose of making improvements to the existing method, it is useful to be able to visualise the problem we are trying to solve. Figure 2 below shows an MRI scan of a mouse the segmentation of whose ventricle that I will use as an example throughout this report. In figures 3 and 4 are plotted the profiles of two typical line cuts through the ventricle. These outline the two main problems of Live-Wire that we are trying to address.

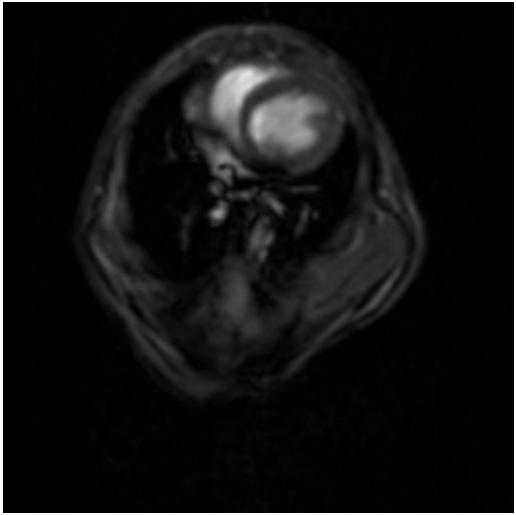

In the first plot, the boundary of interest shows characteristic edge features and a strong gradient. However, note the close proximity of a stronger boundary that will have the effect of attracting the live-wire. This example suggests that provision for the Live-Wire to preferably trace boundaries with certain features would be useful. On-the-fly Live-Wire training, presented in Section 2.3.4 provides a powerful method for solving this problem. Improvements presented in Sections 2.3.2 and 2.3.3 also have a beneficial impact to it.

Figure 2.
MRI scan of a mouse body.

The second plot, on the other hand, shows a gradual, unclear boundary of interest with several close points of Laplacian zero crossing and similar gradients. Hence, we can expect the result of this segmentation to be rather inaccurate and inconsistent, in addition requiring more user effort. The live-wire direction feature function, presented in Section 2.3.2 that favours smooth boundaries, as well as on-the-fly training provide a solution to this problem.

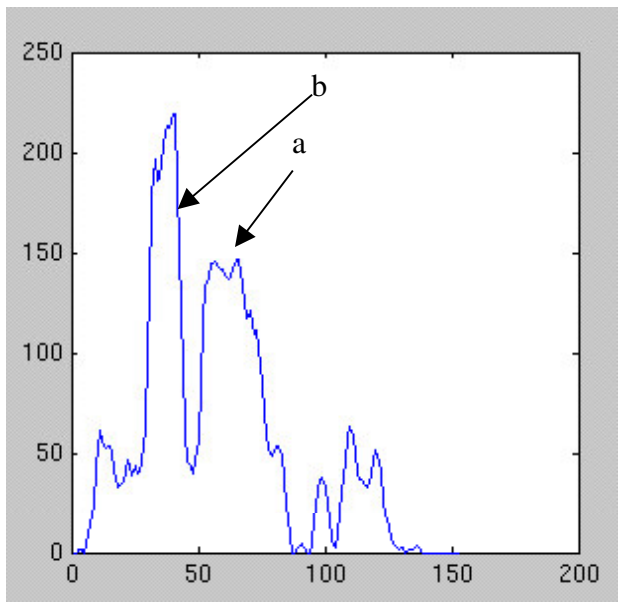

Figure 3.

Greyscale intensity profile of a line cut through the mouse ventricle. (a) Region of interest, (b) stronger, interfering boundary.

Figure 4.

Greyscale intensity profile of the second line cut through the mouse ventricle. (a) Region of interest, (b) gradual boundary with weak boundary properties.

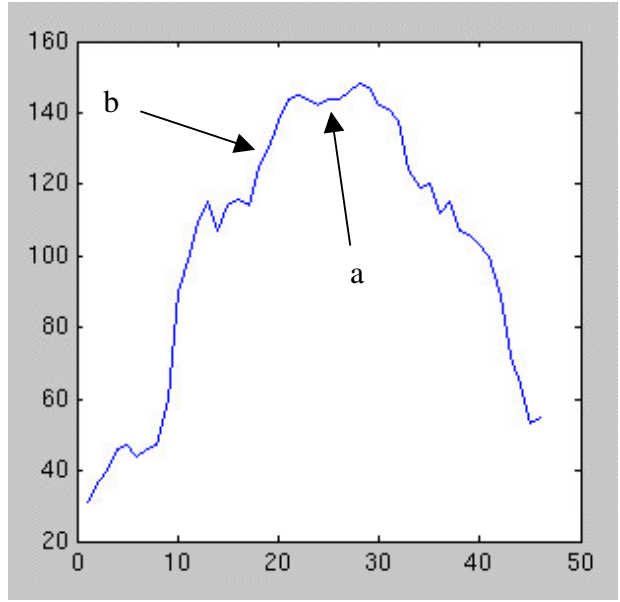

### 2.3.2 Locally optimal feature functions

An important class of improvements to Live-Wire includes various changes and additions to image graph edges calculation. A particularly interesting ones are those that do not conform to the optimality principle (see [3]) behind Dijkstra's algorithm. In particular, my application of Live-Wire 2D is anisotropic in that an edge weight of a pixel depends on how we have arrived at it. Since anatomical objects are smooth and unlikely to have abrupt discontinuities preference is given to the current boundary direction, decreasing as the curvature increases (figure 5). As mentioned before, these feature functions invalidate the use of optimality principle. However, practical results show that the optimal path between two points is still roughly optimal (gives visually good results), but a number of high curvature elements, caused by noise, has been smoothed. In the same class of feature functions is the optimisation proposed by Mortensen and Barrett [1], where the local gradient direction, as opposed to only its magnitude, is used, but I have not used it in my project.

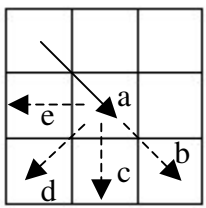

Figure 5.
Direction based feature function weighs edges of the graph giving preference to boundaries with low curvature.
(a) Current direction of the boundary, (b), (c), (d) and (e) directions in order of preference.

One of the main problems that Live-Wire exhibits is the already mentioned distraction coming from neighbouring edges. In attempt to minimise it, I propose a new locally optimal feature function based on the normalised standard deviation of the intensity edge features. This function can be implemented to work very fast requiring only one division and addition per new pixel.

### 2.3.3 Path cooling and path heating

Depending on the size of the object to segment, the Live-Wire method may require a substantial number of seed points for complex boundaries. This not only makes the method more labour intensive, but it can also compromise segmentation accuracy if one's selection of boundary points gets inconsistent. A way of getting round this problem lies in the optimality principle, which in this particular case says that if optimal paths from a seed point to two different points have a point in common, the segment of the path from the seed point to that common point will be the same for them (see figure 6). What this means is that as a user moves the cursor, the initial part of the live-wire will stay constant.

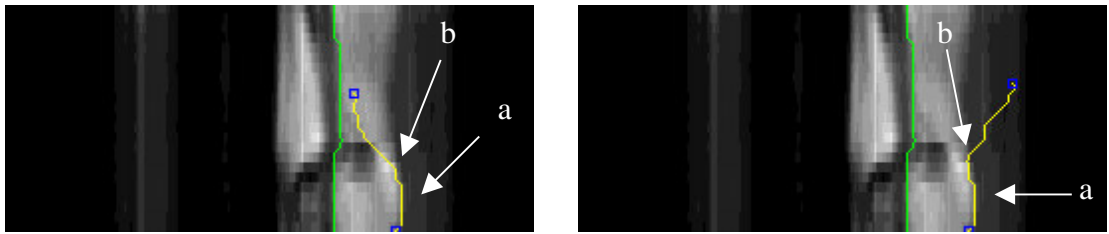

Figure 6.
Live-Wire cooling optimisation automatically generates seed points at the end of invariant segments. (a) invariant initial wire segment, (b) automatically generated seed point.

Exploiting this observation, path cooling saves the user time and effort by automatically recognising a new seed point, as the last point of the live-wire segment that has stayed constant for a certain period of time. The analogy from which the name of this improvement has been derived is that the path cools when it is static until it eventually freezes, generating a new seed point.

The premise behind path cooling is that the initial segment, the one that has most distinctive edge features, is the part of the boundary that we actually want to select. If this is not so, such as in cases of presence of strong interfering edges near the edge of interest (see figure 7), the user is forced to pick short wire segments, making the process very slow. To solve this, in practice very common (see section 6) case, we propose path heating optimisation. The method is opposite to that of path cooling: the wire heats as it remains unchanged, that is, the weight of its edges increases, allowing for strong edges to weaken, when the wire snaps to previously weak ones. The process of path heating can be implemented either by setting a timer that incrementally heats the wire at constant time intervals, or allowing the user to actively control it.

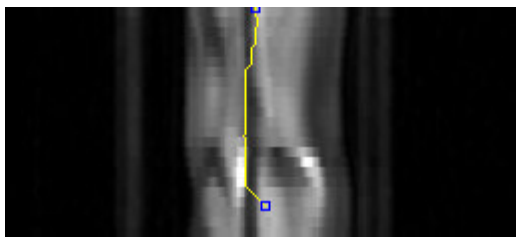

Figure 7.
Strong neighbouring edge interferes with segmentation of the edge of interest.

### 2.3.4 On-the-fly training

Live-Wire on-the-fly training consists of identification of a typical boundary segment strip by the user. If the segment features are indeed typical, this allows to define a mapping from the set of boundary features to estimate of how likely they are to be the part of the boundary that the user wants to segment. This mapping is then used as a function that maps a feature, such as gradient magnitude or pixel colour, to the graph edge weight contribution in the Dijkstra's algorithm. Since the aim of training is to improve accuracy, especially in the case of strong edges near the edge of interest, it is important that the selection of boundary segment is unambiguous. The process of identifying typical boundary segments is done by allowing the user to paint the boundary segment on one of the slices. See figure 8. In the interest of the above mentioned issue, that only one strong boundary is painted, the user is given a set of paint brushes of different widths, with thinner brushes used for more problematic, delicate boundary segments.

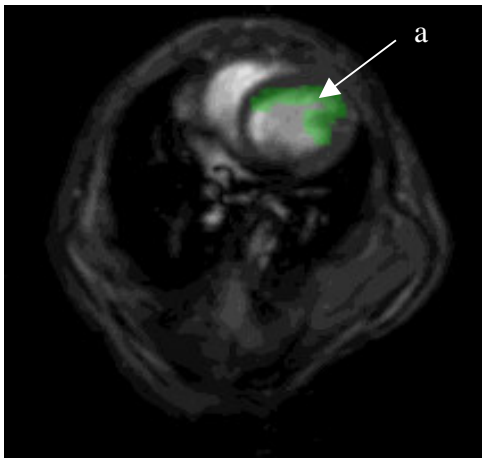

Figure 8.

On the fly Live-Wire 2D training - user uses brushes of different sizes to paint the boundary of interest. (a) semi-transparently painted boundary.

# 3 LIVE-WIRE 3D

## 3.1 Basic extension to 3D

The Live-Wire method can be extended to three dimensions in a number of ways. In the method considered in this report it provides an efficient way of performing Live-Wire 2D segmentation in each of the planes of images through the 3D object - which we assume has been imaged in parallel planes. The efficiency is ensured by a careful compromise between accuracy, speed and user involvement in this process.

In our version of Live-Wire 3D, the user first identifies, what we call *segments of constant topology*. Intuitively, these are volumes between pairs of slices, such that the contours in the slices' planes do not change 'much'. Formally, the requirement is that within a segment of constant topology, the boundary changes of interest are smooth in the sense that none of the boundaries disappears or breaks from slice to slice. After segments of constant topology have been chosen, 3D segmentation is performed in each.

The next step consists of creating a series of orthogonal cuts through the 3D data set (figure 9) and prompting the user to perform Live-Wire 2D segmentation in each (figure 10). This allows for automatic evaluation of two points per orthogonal cut in the slice planes.

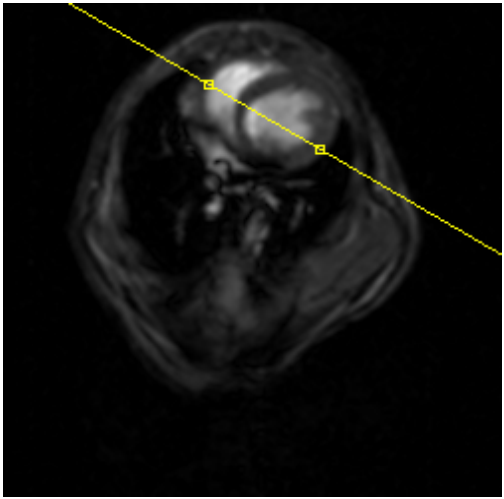

Figure 9.

User is creating an orthogonal cut by selecting two points on a plane of a data slice. Line defined by these two points defined the intersection of the slice and orthogonal plane.

What this allows us to do is to gather enough information about the boundary in each slice plane, by making a small number of 2D segmentations. If the cuts are strategically chosen this allows for accurate automatic application of Live-Wire 2D, with the points in the planes of slices as consecutive seed points.

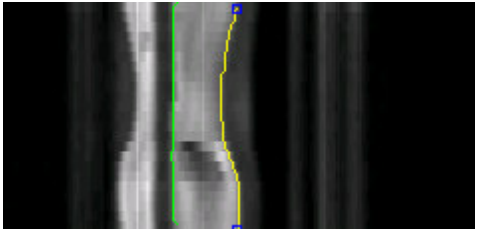

Figure 10.

2D segmentation of an orthogonal cut created by the user.

## 3.2 Improvements to Live-Wire 3D

In the basic extension of the Live-Wire method to 3D we do not use boundary information calculated for one slice, as an a priori information for the segmentation in the subsequent slices. Since the object of interest is most likely smooth, we know that boundaries in the neighbouring slices are likely to be similar to each other. This knowledge, however, should be easily incorporated in the existing 2D method.

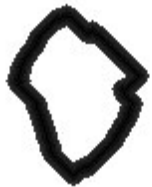

One use of this information is suggested by Schenk et al [12]. They propose the use of a distance transform restriction of effective image area. It consists of creating a strip around the boundary of the previous slice, and restricting the pixels forming the search graph to those belonging to it (see Figure 11). This has both the effect of improving the accuracy of the 3D segmentation, and dramatically improving the speed of calculation as the graph size is reduced.

Figure 11.

Strip around the bondary of the previous slice. Dijkstra's search in the next slice is performed in this area only.

A distance transform represents a matrix of the same size as the original image. It is initialised to a high value (say $3*image\_width$) for non-boundary and 0 for boundary pixels. This represents an initial lack of knowledge about the distance of a particular pixel from the boundary, apart from those exactly belonging to it. Then two chamfering processes are applied (see [12] for further information), using the pattern given in the figure 12. In each pass, the pixel considered is the one below the middle element of the pattern. The first pass is performed from top to bottom, and from left to right, while the second one is bottom up, right to left. In both cases, for each non-empty cell in the pattern, a sum of the pixel below it and the value in the cell is calculated. The value of the pixel in consideration is then assigned the minimal of these.

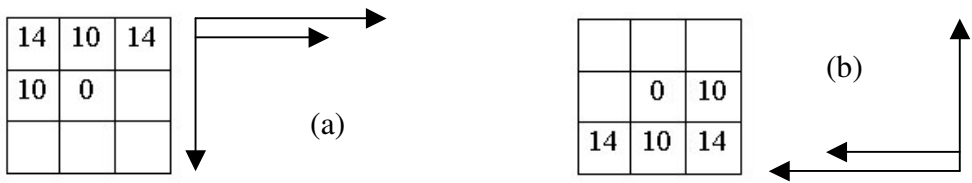

Figure 12. Chamfering patterns and the directions of their application. (a) Pass 1. (b) Pass 2.

Essentially, this is a dynamic programming approach to calculating the distance of a pixel from an arbitrary object. Note the values of the pattern applied. The value of 10 is chosen so that two significant digits are retained in double to integer rounding for diagonal pixels, and represents the unit distance between neighbouring pixels. The value of 14 represents the distance of the diagonal pixels, which is why there is factor of $\sqrt{2}$ in it. Each chamfering process can then be seen as an application of the optimality principle. The minimal distance of a pixel from the boundary is the minimal distance to another pixel summed with the minimal distance between the two pixels.

After the distance transform of the boundary of a previous slice is calculated, it is thresholded and the region below the threshold is used to restrict the Live-Wire 2D search. This, however, leaves us with the issue of choosing the threshold value. Surprisingly, none of the papers that I have seen on Live-Wire 3D (see the list of references) discusses this issue. The solution I came up with was to produce an estimate of the required stripe width. This width should be dependent on the rate of change of boundary through the third dimension since for greater boundary changes there is more uncertainty about the new predicted position. For this reason, I use the knowledge gathered thought the orthogonal slices. The minimal width is estimated using backward differencing on the boundary in the orthogonal cuts, and setting it to be the maximal of these. This gives an estimate of the minimal width. This width is scaled by the user adjustable parameter, in the range of $1.1 - 2.0$, that can be seen as a 'safety factor' that ensures inclusion of the real boundary in the search region. I further investigate its impact on the performance of Live-Wire 3D in section 6.

# 4 IMPLEMENTATION ISSUES

A significant part of my project consisted of Live-Wire 3D implementation. The intention was to produce a program to implement this method, so that various aspects of its performance could be evaluated on clinical data. The obvious issue in all interactive tools is that of time efficiency, but a range of unforeseen problems gave raise to the realisation of key issues, and thoughts of possible improving modifications.

Due to vast amount of code in my project and space limitation in the report, here I do not describe the development process in full. I have therefore selected a range of key issues to present. These include a general overview of the program I have developed as well a number of problems I encountered and ways I got round them. My code complies with the standard Java programming patterns, variable naming is suggestive and consistent, and the code is thoroughly commented, so that even the parts out of the scope of this report should be relatively easy to follow through. Please see the javadoc generated documentation on the accompanying disk as well.

## 4.1 Software design

A challenging aspect of the design of my project lied in an open-ended aim at the beginning. This required a robust, layered design and continuous requestioning of validity of my decisions. For this reason, I made sure to:

- make a clear separation between logical units,
- make clear and sufficient comments, and
- keep a log of my thoughts and problems that I had.

As the project progressed, independence between logical units proved to be very important. This was especially evident in the ease of development, and ease of adapting the program to conform with newly set criteria as new features were added. During the whole of my project I maintained a clear separation between computational classes, and classes managing user interactions and GUI. In this vein, computational classes do not rely on any particular visualisation, or even a way of user interaction. They interact with their environment by a set of class-specific primitives intrinsic for their behaviour, making full potential of code reuse.

## 4.1 Project structure

An important thing in development of any modern software is that of smart code modularization. At all time I had to think about logical code separation, so as to make it as easy to develop, but also as robust as possible to account for future additions, some of which I might be unaware at the time. To give an overview of the global project structure, mainly the role of Java classes that I developed, I present a concise description of each, grouped into logical units.

### 4.1.1 General purpose classes

- Static class *Debug*

As the name suggests, this class has been developed for debugging purposes. Its main functionality is in displaying messages with a precise time stamp and type, which is of significant help, especially in multithreaded applications. In addition, it allows for an easy way of turning some messages off, or indeed all, for the final version of the program.

- Static class *PolygonMath*

The Java API, even including the *vecmath.jar* package, is weak in polygon geometry manipulation. Extraction of orthogonal slices through the 3D data set, or efficient restriction of filters to the area of interest, to name a few tasks where I needed it, suggested a need for creation of this class. In addition, looking forward at the time, I recognised that I would probably need such functionality in 3D-mesh creation around the segmented object.

- Class *PolygonTriangulator*

Often, when handling polygonal data, and especially so for non-convex polygonal regions, we are interested in handing it in a triangle by triangle fashion. This requires 2D-mesh creation over a polygon. Class *PolygonTriangulator* implements one way of mesh creation, and can iterate through the mesh returning one mesh triangle at the time for processing. Mesh creation as described above is not unique, and usually one can formulate a fitness function that expresses how good a certain mesh is. For example, a popular optimal mesh requirement is the greatest minimal angle meshing. However, in my application, I did not need such sophisticated meshes, so I settled with a mesh that does not necessarily satisfy strict optimality criteria in the context I have used it in.

- Static class *ImageUtils*

This static class is a library of various general image-processing tools that I needed. It is used extensively in many places of my code. Some of the functions that it implements are RGB to intensity matrix conversion and vice versa, extraction of a line of pixel values, numerous filters etc.

- Abstract class *Filter*

One of the goals of my project was to explore possible modifications or extensions to Live-Wire for performance improvement on MRI data. This particular imaging method is recognised to suffer from a range of different distortions. Filtering of these images is further complicated by the properties they exhibit. Non-linear behaviour, tissue specific distortions and anisotropic distortions are just some issues that we are dealing with (see [11]). For this reason, I made a decision to investigate the effects of some common filters on the Live-Wire segmentation process. The abstract class Filter specifies some standard filter properties.

**4.1.2 GUI classes**

- Classes *DoubleField* and *IntegerField*

At the time of implementing my code, the highest Java 2 SDK version that was available was 1.3.0_1. Unfortunately, a GUI component such as *JSpinner.NumberEditor*, that imposes restriction on the text entered, available in later Java 2 SDK versions, is not available in this one. For this reason I needed to implement the above two classes. The former implements a text field that accepts a double precision floating point number from a specified range, while the latter does the same for integers.

- Classes *AnisotropicDiffusionDialog, ContrastDialog, HistogramDialog* and *UnsharpMaskDialog*

These classes implement the GUI dialogs for setting up the parameters for the appropriate filters. The user is allowed to manually change the default values of the filter parameters, restrict the filtering, where appropriate, to a polygonal region and preview the effects of these changes.

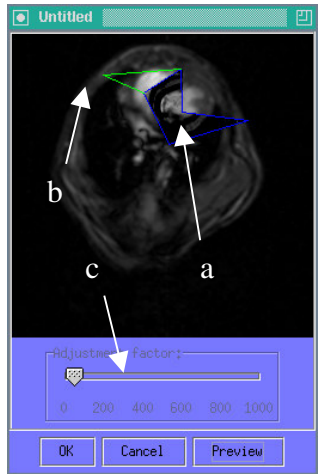

Figure 13.
*UnsharpMaskDialog* class provides a graphical dialog for previewing the application of unsharp mask filtering. (a) Polygonal region to which the filter is applied, (b) mouse click adds a new polygon vertex, (c) a slider allows convenient setting of the parameter.

- Classes *ImagePanel* and *PaintingImagePanel*

Class *ImagePanel* implements a panel that displays a specified static image. It is used in the filter dialogs, when the original image and/or preview display is needed. *PaintingImagePanel* is implemented for the purpose of on-the-fly training. It displays an image, with the addition of the user being able to "paint" on it by pressing and moving the mouse. It also provides means of other classes obtaining information about the painted area.

- Class *OptionsDialog*

This is a dialog for setting up various parameters for Live-Wire 2D and 3D. Amongst others, these include the size of the buffer for storing MRI slices, the safety factor for the distance transform and settings for the heating and cooling optimisations to Live-Wire 2D.

- Class *TrainingDialog*

This class presents the user with a dialog for on-the-fly training of Live-Wire 2D. The paradigm is the same as described in section 2.3.4. The user uses different brush sizes to paint typical boundaries. The information about the painted regions is the processed (see class *FeatureFunction*) to account for detected features of the boundary. The user is allowed to preview the relative weighing of the image pixels, as a greyscale image, by clicking the "View" button. Depressing the "View" button switches to painting mode again, and the user can modify his selection accordingly until the desired result is achieved.

- Class *LiveWire2DDialog*

This dialog is used for all user-driven Live-Wire 2D segmentations. The user is displayed a *LiveWire2DImage* object that allows Live-Wire 2D segmentation, and after the segmentation process is finished the dialog offers the button that accepts it. In addition, the user is allowed to select whether and which optimisation is to be used. Two allowed optimisations are wire cooling and wire heating optimisations (see section 2.3.3).

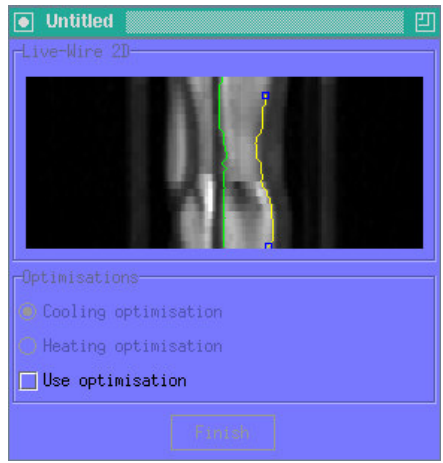

Figure 14.
Live-Wire 2D segmentation process displayed by a *LiveWire2DDialog* class.
(a) Easy to turn on optimisatons.

- Class *SliceNavigator*

*SliceNavigator* is used to quickly navigate through the set of object slices. It is a panel that displays horizontal lines, one for each slice, different segments coloured differently, and allows the user to select one of them by clicking it. Its use in my application is shown in figure 15.

- Class *LiveWire2Dimage*

*LiveWire2DImage* is a panel that displays the Live-Wire segmentation process interactively, by means of interacting with the underlying *LiveWire2D* object. On every boundary event, it redraws the updated information. This includes the currently created boundary segment in one colour, as well as the current live-wire segment in another. For example, cursor movement, if a seed point exists, results in the changing the end point of the live-wire.

- Class *LiveWire3D*

LiveWire3D is the top-level class that implements the graphical user interface (GUI). Its main role is to manage the layout of graphical components and delegate the actions to appropriate classes when user action is performed. See figure 15.

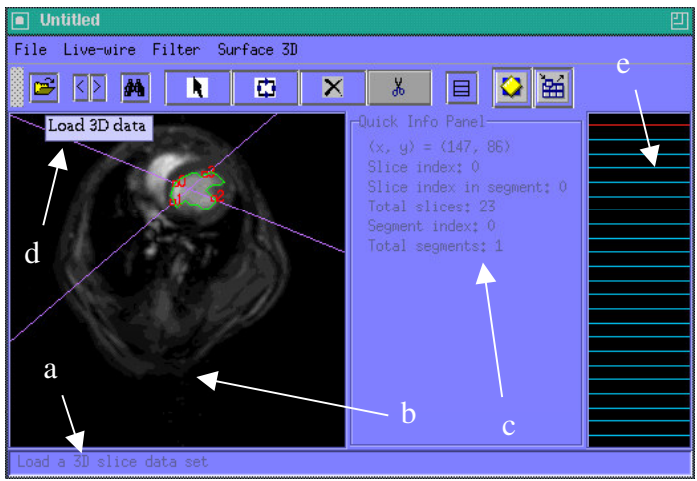

Figure 15. Main screen of the application. (a) Explanation of use of the component beneath the cursor, (b) *SliceBrowser* displays the current slice, orthogonal cuts and the segmented bondary, (c) information panel, (d) concise action description, (e) *SliceNavigator* allows for quick choice of the data slice.

### 4.1.3 Analysis classes

- Classes *AnisotropicDiffusionFilter, ContrastFilter, HistogramFilter* and *UnsharpMaskFilter*

These classes respectively implement anisotropic diffusion, contrast enhancement, histogram equalisation and unsharp-mask enhancement on rectangular matrices of pixel values. All of them are used in the pre-processing stage, before Live-Wire boundary extraction. I also considered an automatic way of detecting potentially beneficial automatic application of these, for reduction of user involvement and speed increase.

- Class *BoundaryEvent* and interface *BoundaryListener*

These cross a gap between a *LiveWire2D* object and classes monitoring the segmentation progress. This is in most cases the GUI, in the interest of updating the displayed information. The *BoundaryEvent* class encapsulates events that happen at various stages of Live-Wire 2D segmentation, when a live-wire is updated i.e. when the cursor has moved, when a new segment is added to the boundary, or when the segmentation has finished. The *BoundaryListener* interface defines actions that an object registering itself to listen to boundary evens need to possess.

- Class *FeatureFunction*

FeatureFunction represents a mapping of a certain feature, represented by an integer in the range 0 - 255, to a relative feature weighing in the range 0.0 - 1.0. This class, once created, is used in Live-Wire 2D for weighting features of image pixels for graph edge calculation.

- Class *LiveWire2D*

The *LiveWire2D* class is the class implementing the behaviour of the 2D segmentation method. It consists of a public class that handles interface to the outer world, such as dispatching boundary events to the appropriate listeners. The computation of boundary segments and the live-wire is done by a protected class that runs in a separate thread. This design decision was motivated by real-time requirements. Dijkstra's algorithm execution happens in parallel with user actions, and in particular user interaction, which allows for most of the computation to be performed before its result is needed. The design and implementation of this class is further described in the section 4.2.1.

- Class *SliceBrowser*

This class implements a browser through the 3D data set - a panel that displays one 2D slice and allows its environment to view the next, the previous slice, or select an arbitrary one. It handles the behind-the-scenes caching of slices, loading of data from files, scrolling and zooming of images and a number of other utility methods (see javadoc generated API documentation for example).

## 4.2 Design and implementation highlights

### 4.2.1 *LiveWire2D* implementation and design

The Live-Wire method in two dimensions forms the basis of the 3D segmentation I have implemented. As such, this was the first class that I developed in this project and it is worth spending some time on the description of its implementation and structure.

Following the general design strategy I have described in the section 4.1, the first step in the design of class *LiveWire2D* was to formulate a clear specification for it. This specification had to include behavioural specification i.e. how the class interacts with other software components, as well as a performance specification, since as in any real-time software, performance is of critical importance.

The former was relatively clear in this case; the behaviour that we want a *LiveWire2D* object to have is to be able to accept a new seed point or a live-wire end point. Note that when these requests are issued is completely up to the environment. It can be initiated by a user click, as when an orthogonal slice is segmented, or indeed can be programmed to occur when automatic segmentation is performed by the 3D method in the planes of all slices. Behavioural specification also required the interested environment to be notified of the important events in the segmentation process. Since the environment may be multithreaded, just as the *LiveWire2D* class, the natural way to implement this interaction was via event invocation and event listeners. For this reason, the class *BoundaryEvent*, and the interface *BoundaryListener* were developed.

While behavioural specification influenced and constrained class interfaces to the world, performance specification affected its internal implementation details. As already mentioned, the real-time nature of this class required it to be threaded. However, for the sake of code reuse, from the standpoint of somebody who wishes to use this class, treating it as a thread requires handling too much detail, and unnecessarily reveals its implementation specifics. For this reason, I did not make the *LiveWire2D* class extend *java.lang.Thread*, but rather made it a thin layer of interface between the main computing thread, *DijkstraThread* class, and the environment. This hides the implementation specific details and makes the API easier to use due to a higher level of abstraction.

Turning to the implementation of *DijkstraThread*, the first place to look for potential speed up was Dijkstra's algorithm. As with all algorithms, the data structure we choose to work with can significantly alter its running time. This is very pronounced in my project, as the graph size I was dealing with was typically of 80.000 to 500.000 nodes. This seems like a rather big number, but fortunately we can use a priori knowledge about the graph to significantly simplify various operations. Namely, we know the exact topology of the graph, as for each node we know which other nodes are its direct neighbours. Needless to say, I have used this observation when adding new discovered nodes with no need for additional memory storage or computation.

The last stage of Dijkstra's algorithm is the reconstruction of the optimal path. Using the optimality principle, this path can be calculated in reverse, starting from the end point until the seed point is found. Alternatively, an auxiliary result can be stored during the computation of the optimal path, making its reconstruction straightforward. The choice of preference between two approaches, of course, depends on a particular application, due to the trade-off between memory use and computation time. Since in modern computers 80 - 500KB of memory is not very critical, my decision was to use the latter method, as the speed of computation has shown to be more critical and harder to achieve.

As with most threaded applications, issues regarding synchronisation are always there. When I started with the design of the *DijkstraThread* class, I quickly realised that a rush of new seed points, such that could occur if a user clicked quickly several times, should not cause any problems. For this reason, I keep a buffer of incoming requests, such as live-wire movements, seed point creation etc. and handle them one at the time. This decision very much influenced the design of the main loop in the run method of the thread. To ensure means of safe thread termination, the while loop is guarded by a check whether it is still valid i.e. whether the thread is to be shut down or not (refer to code excerpt 1). At the beginning of this loop the thread waits for new incoming events. If no events are present it sleeps until it is awaken. Then it checks (strictly, this is not necessary, but it is a good practice to do so) whether the queue of incoming requests is not non-empty in which case it starts processing the first one.

```
while (threadActive)
{
  while (bufferEmpty)
   waitUntilNotified();

  { .. Get new request from the buffer .. }
  { .. Initialise Dijkstra's algorithm .. }

  while (threadActive  &&  valuesValid  &&  moreNodesLeft)
  {
    extractMinDistanceNode();
    addNodeNeighbours();

    if (interesteInWireToTheNode)
     reconstructPath();
  }

  if (threadActive  &&  valuesValid)
   { .. The complete graph search is done .. }
}
```

Code excerpt 1.
Simplified pseudo-code for the main loop of *DijsktraThread* class.

This comprises the inner loop of Dijkstra's algorithm, in which in each iteration the reached front of nodes is expanded and a shortest path to a new node is calculated. Since we know what the neighbours of each pixel are a priori and there are always at most 8 of them, addition of the nodes in the expanded front is done in O(1) time. If the shortest path to the node of the next request in the queue is found, it is immediately reconstructed. If this corresponded to the placing of a new seed point, Dijkstra's algorithm ends, and a new iteration of the main loop is performed, avoiding wasting of computing power. An additional speed-up and improvement it offered by the methods called when a new seed or destination point is placed. If the flags set up for that purpose indicate that the result of this request is already available, it is immediately provided, with no request actually being placed in the request buffer.

### 4.2.2 *SliceBrowser* implementation and design

Class *SliceBrowser* implements most of the Live-Wire 3D functionality. Since all the quantitatively heavy work is performed by the Live-Wire 2D method, in this class there were not as many places for performance improvement as in *LiveWire2D*. However, performance issues started to arise when I started experimentation with pre-processing of data, using various filters. This was especially the case for anisotropic filters, which are inseparable in multiple dimensions. Speed-up in the stage of automatic contour calculation in the planes of all slices is clearly not possible to improve, as the minimal path search potentially includes the whole graph i.e. image. The place where I realised improvement could be made was when orthogonal cuts were created. I observed that filters have a 'domain of influence' i.e. that only a region of the original image influences the value of a certain pixel after filtering. Orthogonal cuts, as implemented in my project, are created using linear interpolation between pixels intersecting the line of an orthogonal cut in the plane of a slice. The number of such pixels is $O(\sqrt{n})$, where $n$ is the number of image pixels, and filtering of the whole image is inefficient in terms of computing power. This motivated the inclusion of *getInfluenceWidth()* method in the abstract class Filter. This method returns the maximal distance (radius) of a filtered pixel from a pixel in the original image that influences its greyscale value. This modification, coupled with polygon manipulation routines I implemented in the class *PolygonMath* and *PolygonTriangulator*, allowed the filtering process to occur only in a strip along the line of interest. This proved to provide a very substantial increase in speed during the orthogonal cut creation.

Two problems, one more subtle than the other, arose when I faced the extraction of useful information from the orthogonal cuts. The simpler one was to properly reconstruct the order of the points on the boundary in the planes of each slice. This is, of course, impossible without imposing any restrictions on the way orthogonal cuts are made. As a solution intuitive for users, and easy to handle in the code, the restriction on the orthogonal cuts that needs to be obeyed for valid segmentation to take place, is that they form a strictly clockwise, or strictly anticlockwise sequence. In other words, the ordering of the points is such that all the beginning points of the cuts come in sequence one after the other, followed by the end points of the cuts, again in the same order in which they were created. A valid and an invalid situation are illustrated in figure 16.

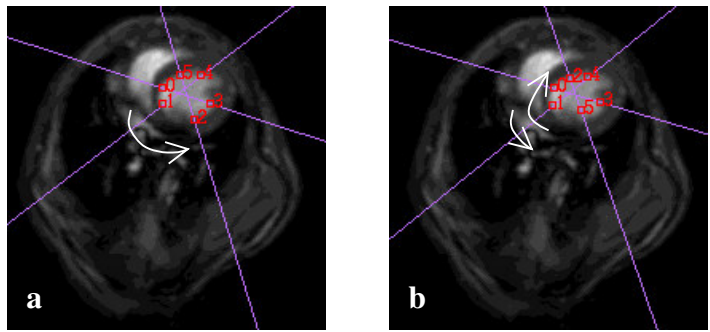

Figure 16.
(a) A valid and (b) an invalid way of creating orthogonal cuts. The latter shows inconsistency in the orientation in which the cuts are specified.

The second problem lied in the choice of points in the data slice plane from the contour in the orthogonal cut plane. It can be shown that in each segment of constant topology, as defined in section 3, any orthogonal slice will not have 'wiggles' that go up and then down. That implies that every orthogonal cut produces two and only two, points in each slice plane. However, although the object might not exhibit such behaviour, the noise content of the images might artificially cause Live-Wire 2D to snap in such a way, leaving us with more than 2 points in a plane for which we have no knowledge of ordering. Observe that the nature of Dijkstra's algorithm implies that the problematic noise in this case is that of high frequency around the point on interest. Using this observation, I solved the problem by taking only one of the multiple intersecting points caused by wiggling, as they are likely to be very close to each other. Indeed, the tests in practice confirm this to be the case, and my solution performed without a failure in the tests we have done.

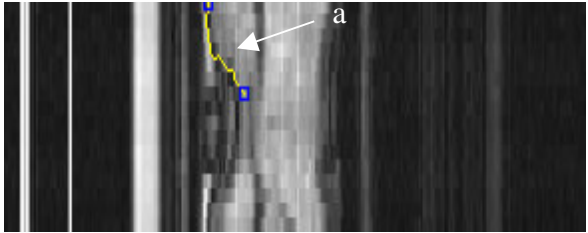

Figure 17.
Orthogonal cut through noisy human cardiac cycle MRI data. (a) Live-wire follows the noise, causing ambiguity in the choice of points in the slices' planes.

### 4.2.3 Object reconstruction and triangular surface mesh generation

The generalisation of Live-Wire method to three dimensions that I have presented produces as a segmentation result a set of contours in parallel planes. For most practical purposes this is not a convenient representation of a 3D object to work with. Most commonly, a continuous, piecewise linear, polygonal representation (a polytope) is needed. Recognising the practical importance of this step, the last task in my project was implementation of object reconstruction as a triangular mesh from a set of contours.

Since the gap between the contours represents a region about which there is no information, it was essential to recognise a set of possible transformations a contour might undergo from one slice to another. This requires certain assumptions to be made in order to formulate a simplified model of the transformations. The model I have used assumes an arbitrary affine transformation between two adjacent slices, having an unknown, but 'small' shape change. Small in this context means that each point of the initial contour maps to the point on the new contour that is closest to it, or sufficiently close to produce visually good results. In practice, this assumption has proved to be valid, since the triangular meshes produced were visually very good approximations to the smooth objects segmented.

The assumption made suggested the following algorithm. First, a set of points along the topmost contour is generated, with equal arc lengths separating them. Starting from one of the points a search for the closest point on the adjacent contour (after application of an affine transformation) is performed along a fixed arc length, this length being a variable parameter. Search for the closest points to the other points of the top contour is then continued from the newly found point. The two resulting sets of points of the same cardinality represent an approximation to the two contours, which allows for easy triangular mesh generation. Each quadrilateral formed by two adjacent points and their closest points is simply divided into two triangles. This process is then repeated with the points obtained on the second contour as initial points in the segmentation of the gap between it and the next slice. The affine transformation applied to the second of the two contours in consideration at any one point is chosen in such a way so as to result in a contour of the same circumference as the previous one, and with an aligned centroid. Consequently, the points obtained on this contour are transformed with the inverse transformation when the triangles are generated.

The choice of the initial vertex in the first contour obviously impacts the overall process by restricting the arc of search for corresponding points. Experimental results have show that the best results are obtained when a point of a convex region of the first polygon is chosen. Figure 18 shows the result of application of this meshing method on a cardiac cycle of mouse MRI data.

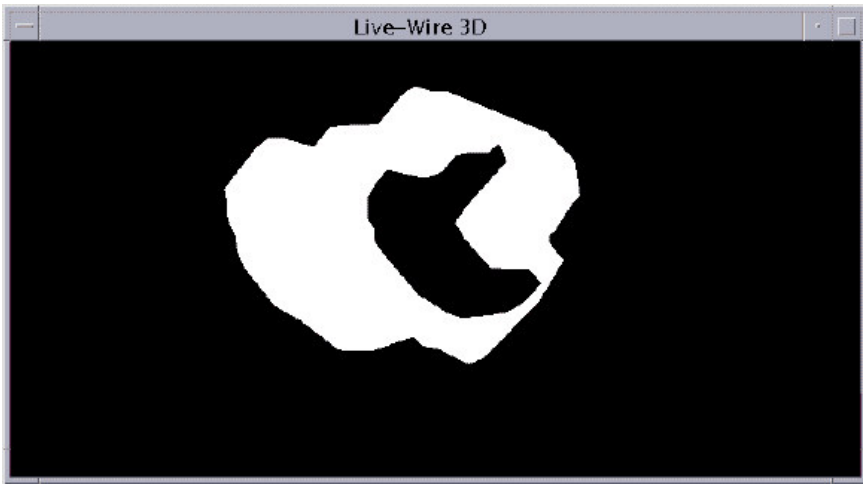

Figure 18a.
3D visualisation of a piece-wise triangular mesh reconstruction of a mouse cardiac cycle (first 15% of the period). No lights have been defined in the scene, so there are no shadows on the object.

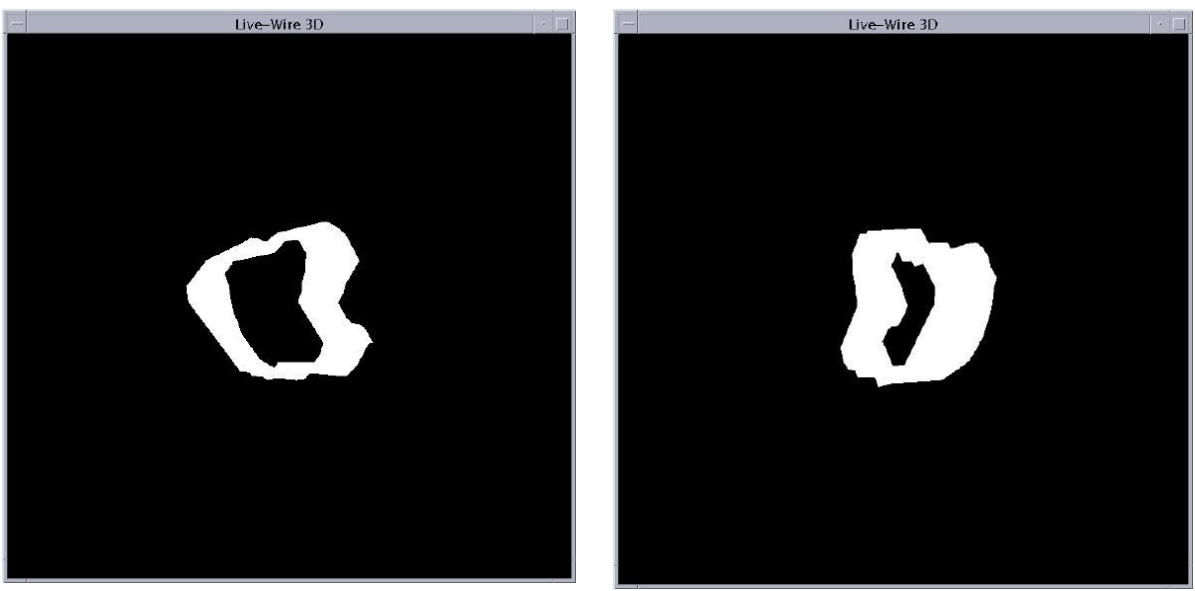

Figure 18b.
Two frames from an animation showing a rotating 3D reconstruction of a mouse left ventricle through time.

# 5 Live-Wire 3D evaluation on MRI data.

Testing and evaluation of the implemented features of Live-Wire 3D was performed on a number of 2D+time MRI data. Three main criteria were used to assess the goodness of the method - accuracy, repeatability and efficiency. Accuracy of a segmentation method refers to how well the result agrees with the truth. The main problem in assessing accuracy lies in the difficulty of establishing the ground truth. For this reason, manually segmented contour by a qualified person is used as the result to compare against. Repeatability is intrinsic for semi-automatic methods, and a very important feature to consider. It refers to the variation in the segmentation result due to manual input of the operator. Obviously, methods with high repeatability (and hence low variation in the results) is to be preferred. Efficiency refers to the practicability of the method with respect to computational time delays.

All tests were performed on a Sun Blade 1000 with clock frequency of 600 MHz and 512 MB of RAM. It is important to note that the application was implemented in Java, and as such interpreted in Java virtual machine, which is obviously not suited for optimal performance. Java was chosen for the following reasons. It has a very wide platform support and has the highest level of platform independence amongst the popular compiled languages. There is also a wide number of supported and well-established libraries for Java, implementing various utility computations. No less important is the structure of Java as a language, which in many ways encourages good programming practice and eases up application development. Deep discussion of this topic is beyond the scope of this report, but sufficient is to mention a very elegant organisation of class inheritance and interfaces, as well as automatic documentation generation using javadoc tool.

## 5.1 Results of segmentation on a mouse cardiac cycle

Complete testing I have performed included applying the method to a number of 3D data sets, but for the sake of ease of comparison, the results I am presenting here are obtained specifically from a MRI mouse cardiac cycle data. The full cycle consists of 23 two-dimensional slices, with temporal resolution of approximately 5.2 ms, corresponding to a mouse heart beat frequency of 8.3 Hz. In this test, I considered segmentation of the left ventricle of the mouse. The first 8 contours resulting from a typical segmentation are show in figure 19.

First, I have explored the basic method without data specific adjustments such as filter application or on-the-fly training. With the aim of establishing quantitative measures, I performed a number of tests, with planned variation of parameters. To make the results more tractable for the reader, I have selected 8 typical (for the set of parameter values) results.

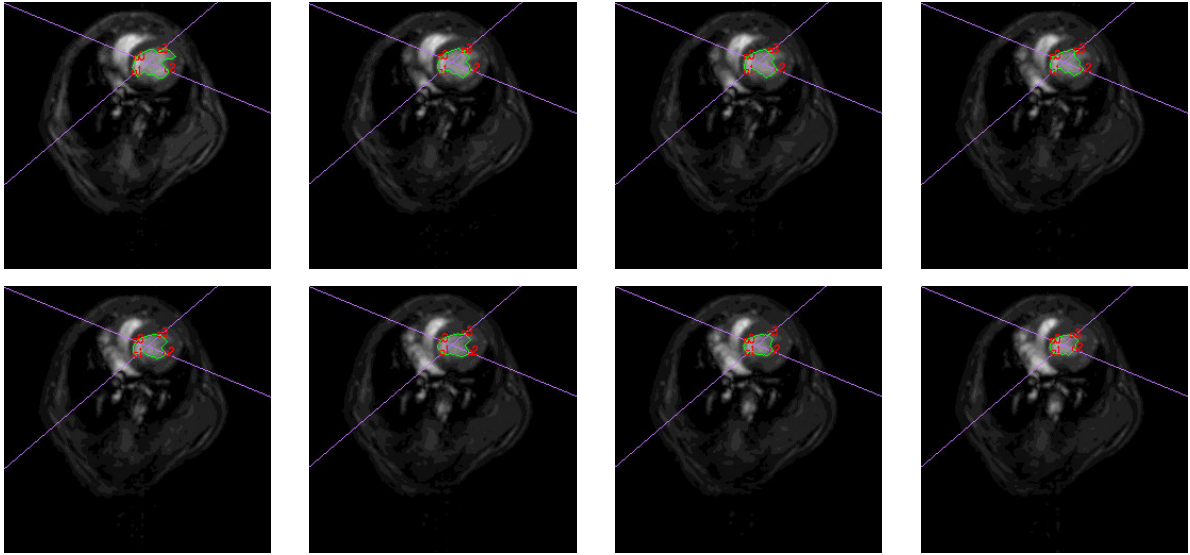

Figure 19.
The result of Live-Wire 3D segmentation in the first 8 slices' planes. Points in red are the seed points for Live-Wire 2D resulting from the two orthogonal cuts, shown in purple, in each plane. Reconstructed contour is displayed in green.

As expected, the number of orthogonal cuts created by the user had the most drastic influence on results. To quantify the variance in the results, I have assumed the error in each slice plan to be symmetrical about the true boundary. With this assumption in mind, as a measure of error between two contours I have used the sum of absolute error as determined from the distance transform of a boundary. In other words, to compare two boundaries, I would take the distance transform of the first one and sum the distances corresponding to all pixels of the second one. Since the error is symmetrical about the real boundary, I have used the mean of all mutual errors as the approximation of error from the true contour. The result for 3D segmentation of the left mouse ventricle resulting from 2 orthogonal cuts is shown in the figure 20. The figure shows the variation of error though the third dimension (time) depending on errors due to manual input, keeping the intended actions the same i.e. the number and intended placement of orthogonal cuts.

An important feature of the presented graph is a peak in error that occurs in the range of slice indexes from 14 to 17. A look at slices in this range reveals the problem. Most pronounced in the slice 14, but also present in slices 15 to 17, is an MRI artefact (Figure 21). Due to its strong edges, live-wire is attracted to it, and a large error occurs. Due to inertia of Live-Wire 3D i.e. due to use of prior information, it takes some time for the contour to recover and snap back to the boundary of interest. Two approaches, within the scope of my project, seemed promising to correct this. The first one was variation, possibly reduction in the safety factor used for the distance transform margin. I have performed a number of tests varying this parameter, but little benefit was perceived, so the results are not included in detail. For small values of the safety factor (close to 1.0), the tendency of the contour to snap to the artefact was lower, but the recovery from the error was slower too.

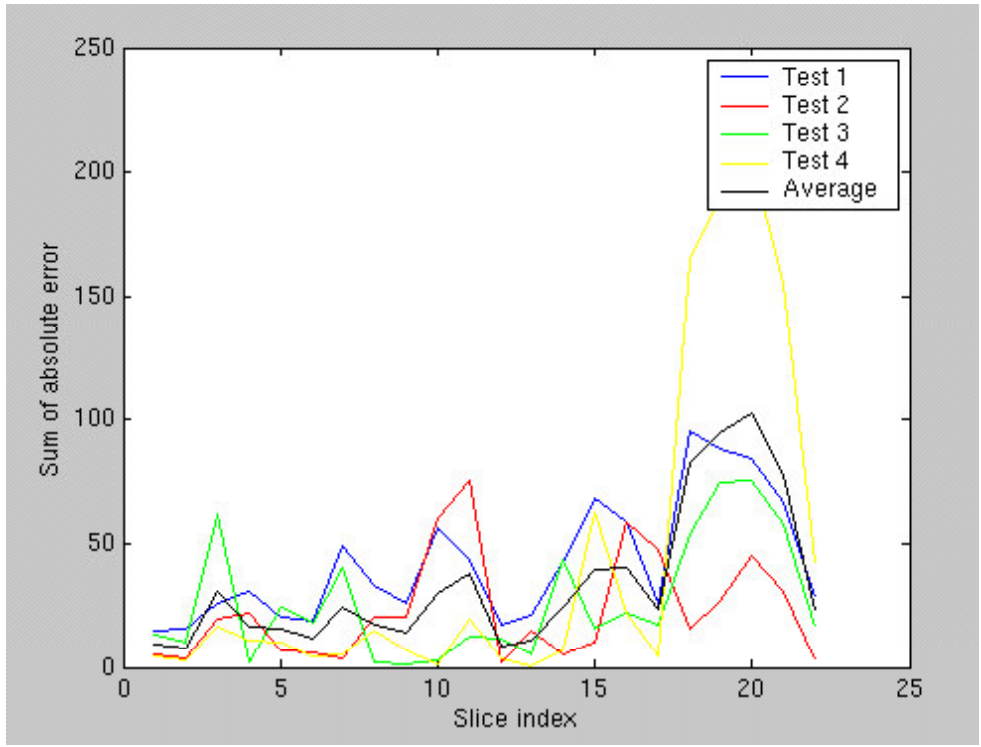

Figure 20.

Error in the contour approximation through time for four test runs, each resulting from two orthogonal cuts, using the best a posteriori estimate of the true result.

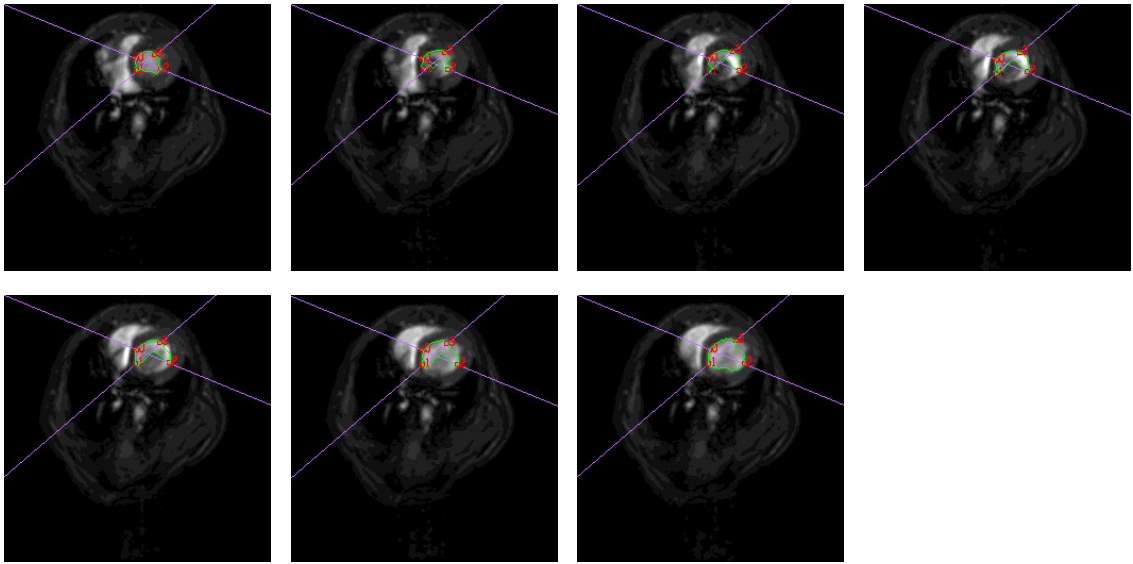

Figure 21.

The result of Live-Wire 3D segmentation in the slices 13 to 19. An MRI artifact with strong edges occuring in the slice 14 attracts the live-wire and produces significant errors. As the artefact disappears subsequently, in slices 18 and 19, the contour snaps back to the edge of interest..

The next solution I tried was to increase the number of orthogonal cuts. Intuitively, this should provide more accurate results, as automatically performed Live-Wire 2D segmentation in each of the slices has more points of constraint. Intuitively much less clear is the issue of repeatability. The results, normalised in the same way as the previous one, is shown in Figure 22.

We see that out hypothesis is confirmed: the absolute error is indeed lower (note different scale) and the accuracy of segmentation is increased. Very importantly, we see that the peak due to artefact in the slices 14 - 19 is significantly reduced. Having established greater accuracy of the latter segmentation, we are in position of more accurately assessing the errors in the former. Figure 23 shows the error of the segmentation with two orthogonal slices, where as the result from the second, more accurate segmentation is used as the ground truth. Of the four results, I have used the one that produced the error vector with the lowest 2-norm, which is shown in green in Figure 22.

To compare *repeatability* of two segmentations, we computed the standard deviation of the error in two cases. Figure 24 shows the variation of standard deviation of the sum of absolute error through the third dimension, superimposed for the two segmentations.

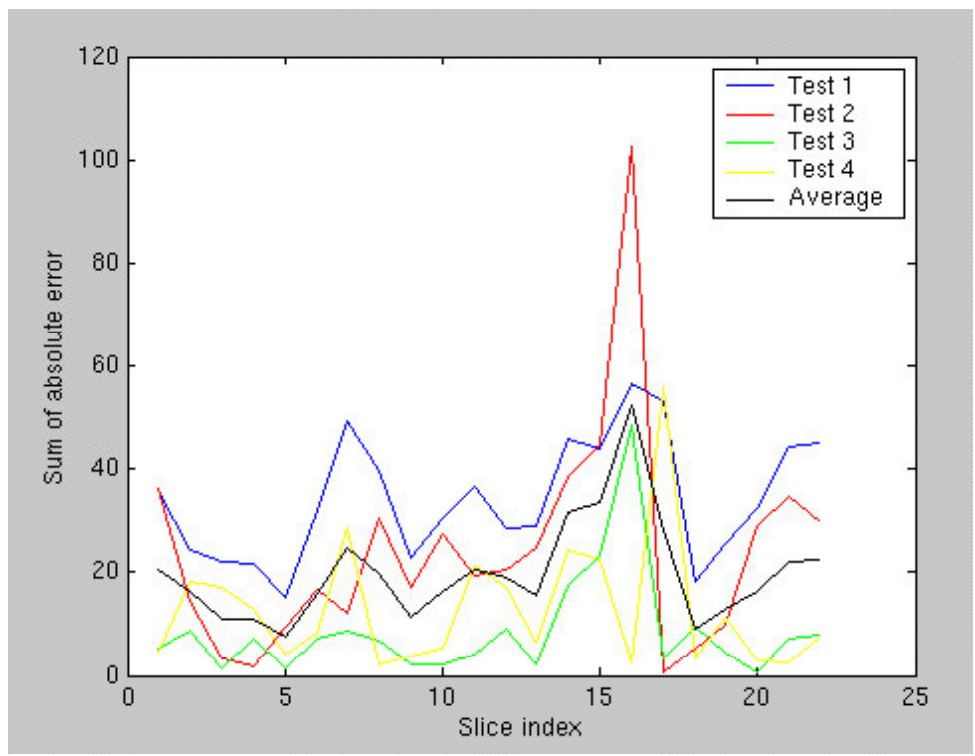

Figure 22.

Error in the contour approximation through time for four test runs, each resulting from three orthogonal cuts, using the best a posteriori estimate of the true result.

From the plot in figure 24, we now clearly see that not it is only more precise, but that segmentation with more (strategically chosen) orthogonal cuts yields more repeatable results. Creation of more orthogonal cuts however, does not require only more time for segmentation, but also more user effort. To put this trade-off into perspective, I created a time profile for a series of segmentations. Table 1 summarises the results for the time and user effort involved in each segmentation.

An interesting observation is that creation of more orthogonal cuts, while requiring more user time (and effort), resulted in faster overall segmentation. The reason for this lies in that the live-wire segments that form a boundary are shorter. Since the front of the nodes reached in Dijkstra's algorithm expands exponentially, this results in a significant

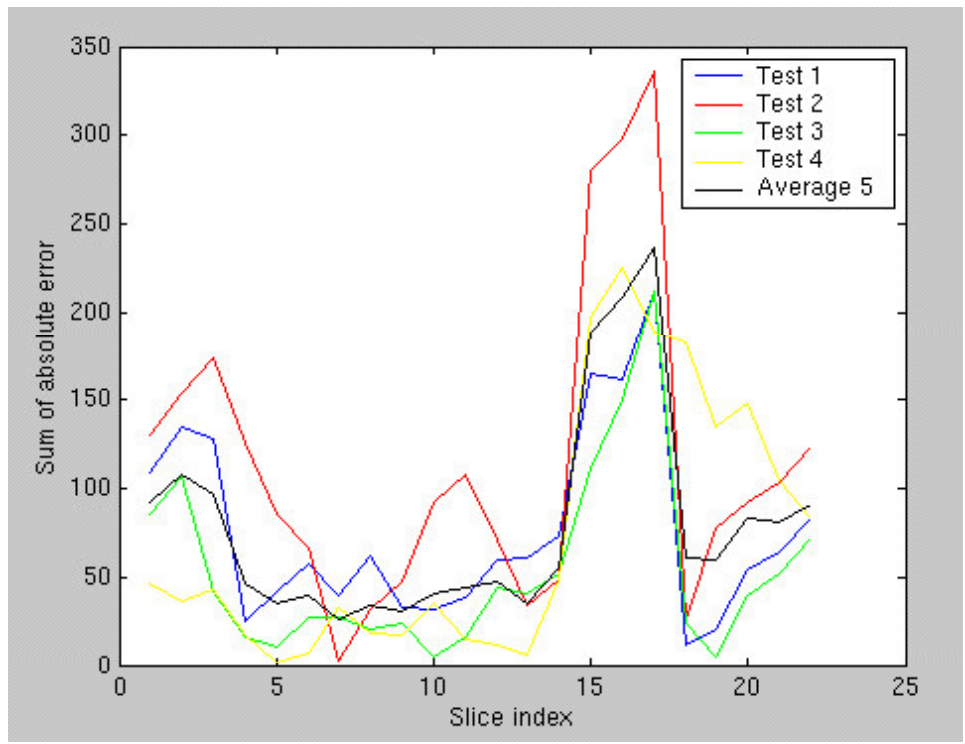

Figure 23.

   Error in the contour approximation through time for four test runs, each
   resulting from two orthogonal cuts, using the estimate of the true result from
   a more accurate segmentation.

reduction in computation time. The conclusion is that there is an optimal number of orthogonal cuts, from the point of minimisation of the total segmentation time. Less cuts than that results in increase of computation time of Dijkstra's algorithm while more results in increase of time spent in orthogonal cut creation and segmentation by the user. Both effects are more pronounced in larger images and greater number of parallel slices.

| Number of orthogonal cuts | Total number of seed points created by the user | User time for Live-Wire 2D segmentation (totals) | User time for selection of orthogonal cuts | Computation time (user not involved) | Total time for 3D segmentation |
|---|---|---|---|---|---|
| 2 | 18 | 55" | 32" | 1' 45" | 3' 19" |
| 2 | 19 | 52" | 32" | 1' 45" | 3' 16" |
| 2 | 18 | 53" | 33" | 1' 45" | 3' 17" |
| 2 | 20 | 52" | 33" | 1' 45" | 3' 17" |
| 3 | 23 | 1' 02" | 47" | 1' 18" | 3' 13" |
| 3 | 23 | 1' 00" | 47" | 1' 17" | 3' 09" |
| 3 | 25 | 1' 03" | 48" | 1' 18" | 3' 13" |
| 3 | 23 | 1' 03" | 47" | 1' 18" | 3' 12" |

Table 1.

Time profile of 3D segmentation of a mouse ventricle for two and three orthogonal cuts.

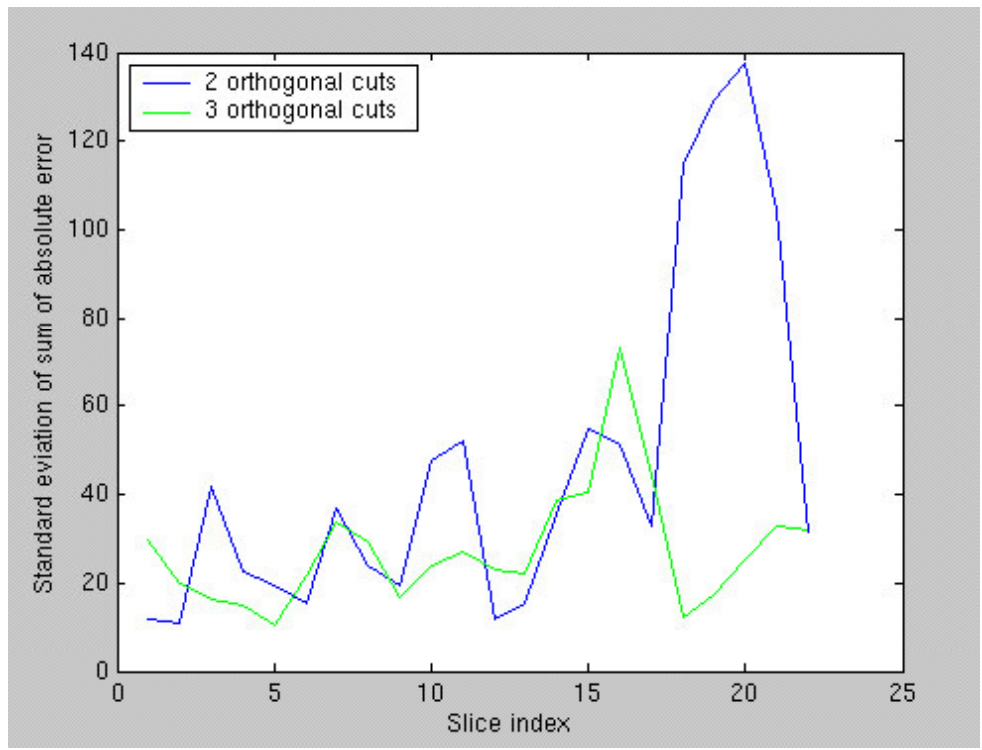

Figure 24.

Comparison between standard deviations of the sum of absolute error through the third dimension for segmentation with 2 and 3 orthogonal slices.

The results of data pre-filtering are much less conclusive. No general behaviour of the method with pre-applied filters was perceived, so I will address performance of each separately.

*Anisotropic diffusion*[2] was found not to influence the results at all. The conclusion is that MRI scans do not contain significant amount of noise compared to the levels of greyscale quantisation and spatial discretisation noise.

*Non-linear contrast enhancement* was found to be very useful at times. With parameters set up correctly, a large amount of features not of interest could be removed and segmentation accuracy was greatly increased. However, an inherent problem is that its effect is very hard to predict on a large number of slices, when only one of them is previewed. This can take a significant user time.

*Unsharp mask filtering* was found to be of very limited use. The main problem with it is in the imperfection of real boundaries, that are smooth, so the filter resulted in distortion (movement) of the real edge position perceived by Live-Wire, and increased interference from strong edges.

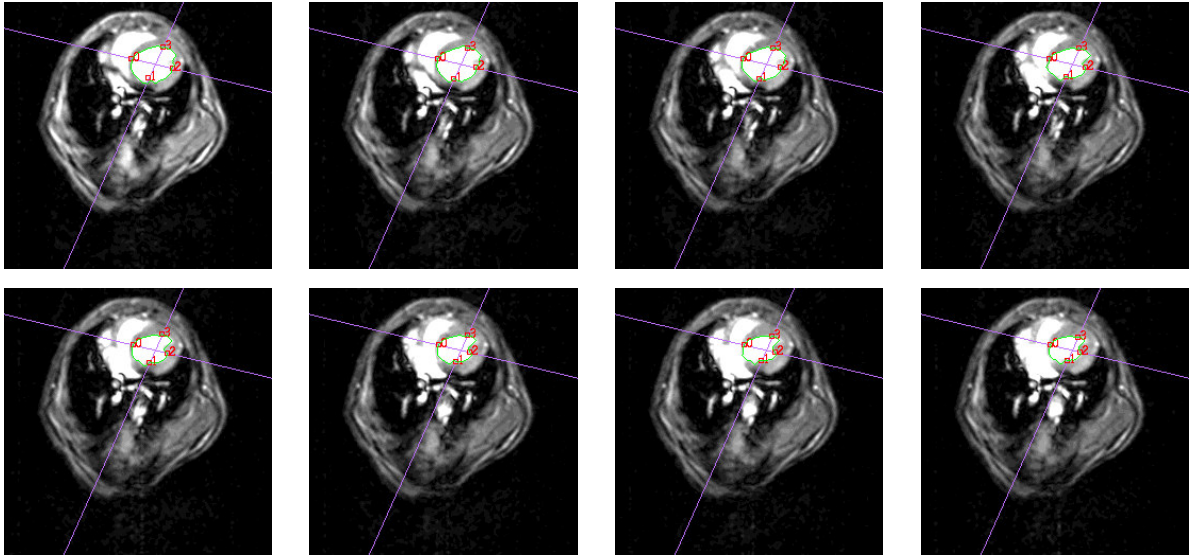

Figure 25.
Non-linear contrast enhancement resulted in an accurate and repeatable segmentation.

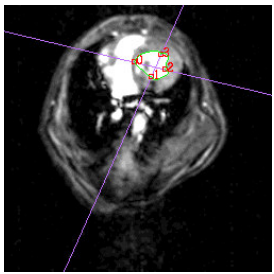

Figure 26.
Fourteenth slice in the 3D data set. Non-linear contrast enhancement vastly reduced the erorr caused by the artefact introduced in this slice.

*Histogram equalisation* was found to amplify noise too much and hence was found not to be very useful. Having in mind that MRI scans are typically very dark, this was to be expected.

The observation that human decision about filter use is very time consuming, and hence defeating the purpose of interactive segmentation, it is natural to ask whether this process can somehow be automated. Approached in this way, this problem turns out to be very difficult. Note, however, that filtering has the effect of changing the image, while keeping the mapping from the pixel features to likelihood of them being edge pixels constant. A different way of doing the same can therefore be to keep the image the same, but changing the mapping - exactly what on-the-fly training does, as described in Section 2.3.4.

The improvement that on-the-fly-training produces critically depends on the way the mapping from a set of features to a set of costs is implemented, and this in itself is a problem on the level of a whole third year project! To remind the reader, recall that the assumption we are making is that the strongest boundary in the painted region is the boundary of interest. Obviously, this is required for the sake of being able to uniquely identify it. The set of functions I experimented with, and that produced rather good results were proportional to a certain power of the gradient and the frequency of occurrence of that feature. The former was chosen due to the assumption stated above - the highest gradients in the painted region are those belonging to the boundary. The latter is a bit more subtle, but easy to understand. It is used to eliminate MRI artefacts, or noise of any type i.e. outliers with big gradients, but low frequency of occurrence, which obviously do not belong to the edge of interest. Experimental results showed the best performance with the function $f = |\nabla I|^3 * frequency^2$, normalised to the working range, and that is the final implementation in my code.

Figure 27 shows a segment of a mouse left ventricle boundary painted to test on-the-fly training.

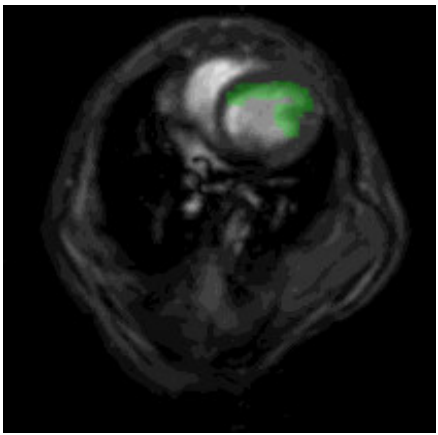

Figure 27.

A typical region of a mouse left ventricle painted by the user, used for the on-the-fly-training of Live-Wire 3D.

---

[2] Frequency in this context denotes the frequency of occurence of a particular feature in a sample.

To assess the effects of the painting on the graph edges' costs reweighting, the user is given the option of visualising the graph edges' costs. The cost corresponding to the image pixels before the training is shown in figure 28.

The improvement is obvious. Note that prior to the training, the major part of the mouse body is shaded in grey with the boundary of interest very slightly pronounced. After training however, the boundary of interests (and of course similar boundaries) turn out much brighter indicating lower cost in Dijkstra's optimal path search. The effects on segmentation are shown below. Figures 30 and 31 show the live-wire between the same two points in the orthogonal cut, without and with Live-Wire training. Without training the live-wire snaps to the stronger neighbouring boundary. After training, live-wire follows the edge of interest.

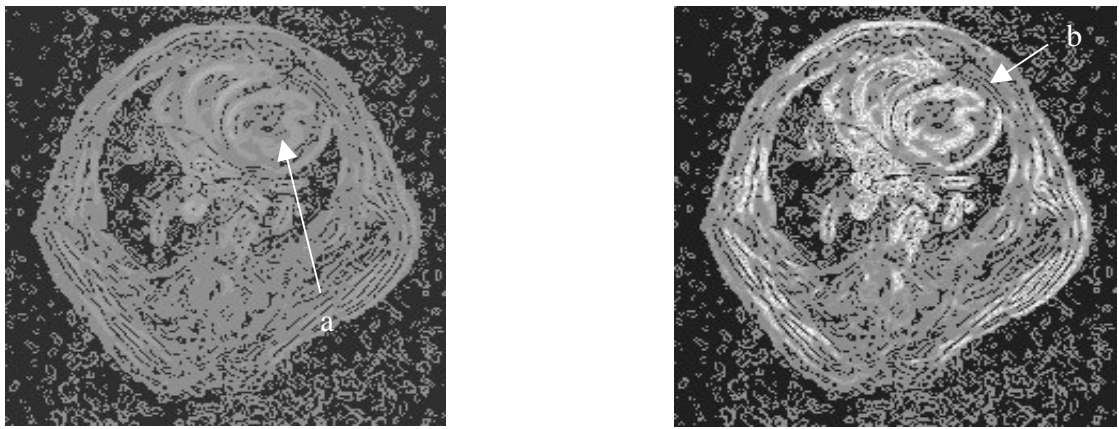

Figures 28. and 29.

Cost corresponding to the image pixels before and after on-the-fly Live-Wire training. Bright pixels denote low, dark ones high edge weights. (a) Initially, the boundary of interest is not very pronounced as seen by the Live-Wire method, (b) training enhances its visibility, and edges of similar properties.

To test the effect of Live-Wire on-the-fly training, we performed a number of 2D segmentations and compared the number of seed points needed for a visually satisfactory boundary detection. The results show that the training reduced the number of seed points

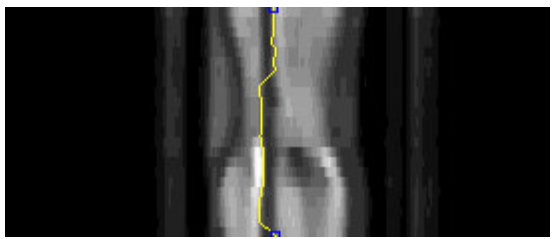

Figure 30.
Before on-the-fly training, the strong boundary near the boundary of interest intereferes with the segmentation – the live-wire is attracted to the stronger boundary.

needed for up 33%. This has the effect of both reducing the user effort, thereby increasing the repeatability of the method and increasing the accuracy for the fixed number of orthogonal cuts. Figures 28 and 29 show the live-wire between the same two points in an orthogonal cut before and after Live-Wire training. Despite the presence of the strong

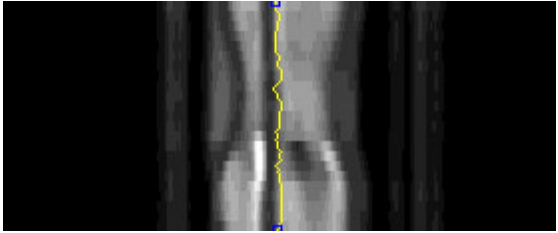

Figure 31.
After on-the-fly training, the graph edges are reweighted and the interference from the neighbouring boundary is reduced.

neighbouring boundary, the live-wire accurately tracks the boundary of interest.

## 6 Conclusion and Future Work

Segmentation of 3D medical images is a challenging task. The Live-Wire method is recognised to be one of the best interactive segmentation methods at the time of writing of this report. The purpose of this project was to design and implement the Live-Wire 3D method for segmentation of 3D images, evaluate its performance on medical MRI scans and suggest and implement improvements to it.

Our contributions are:
1. Implementation using a threaded architecture including classes for GUI and interactions that are not standard in Java API.
2. Investigation of ways to improve robustness and in particular the use of a priori information for 3D segmentation, and improvement to segmentation of boundaries after on-the-fly training. We showed that a reduction of up to 33% in user effort can be achieved.
3. Quantitative evaluation of accuracy, repeatability and time efficiency of the method.
4. Investigation of possible performance improvement using pre-filtering techniques. These had a varying impact on segmentation performance, and we discuss why on-the-fly training provides a better alternative.

In-vivo testing, presented in the chapter 5, has shown performance of the implemented method on a various types of tissues and scans qualities, which highlighted the places for improvement and its behaviour on the data of interest. The segmentation process is intuitive and is performed in a fraction of time required for manual segmentation. Its interactive nature allows the user to view results in real time, correct when necessary. Additionally, a range of tools, such as on-the-fly training allow for automatic adaptation to the data in question. With these results in mind, we confirm that Live-Wire 3D is a very potent segmentation paradigm, and we believe that with development of various tools, some of which have been presented in this report, it will hold one of the leading places amongst the interactive segmentation methods. The tool developed in this project is now being evaluated in a clinical research project.

A problem that still remains is the observed inconsistency and difficulty to analyse the impact on the segmentation that the Live-Wire training has. Since the training was shown to dramatically reduce the necessary amount of interaction with the user, as well as to be a very potent tool for improving accuracy and the speed of the method, future work will concentrate on the choice of the mapping for extraction of salient features of the object edges from the training data provided by the operator.

At the end, I would like to thank my supervisor, Dr. Alison Noble, for her help throughout the project and Professor Andrew Zisserman for most helpful advice on 3D meshing.

# Appendix A - Accompanying Data

This project is accompanied by one floppy disk[3]. The disk contains:

1. full source code of the application,
2. compiled Java byte code, using Java version 1.2.2, Solaris VM (build Solaris_JDK_1.2.2_07a, native threads, sunwjit), archived to a Java readable package using Java Achiever (jar) tool,
3. javadoc tool generated documentation in HTML format[4],
4. project report as a Microsoft Word document, and
5. two textual 2D+time data sets.

All source code files can be found in *oxford/eng/u99oa/year3/project/livewire3d/* and are best viewed using text editor *vi*. The corresponding byte code files can be found in the same directory, but the application is best invoked using 'java -Xmx96m -jar LiveWire3D.jar' command, from the root directory of the disk (e.g. */mnt/floppy/*). Test data can be loaded by choosing an appropriate command from the application menu or toolbar and clicking the file *human.txt*, for human cardiac cycle data, or *mouse.txt* for mouse cardiac cycle, both of which are in */data/* directory. Note that this file can be specified from the command line as well, using the application option *-file*. Further information about the available command line options can be obtained by typing 'java -jar LiveWire3D.jar -help'.

# Appendix B - Software requirements

Software requirements for running the application are:

1. Java Virtual Machine (JVM) 1.2 or later,
2. Java 3D package, with its classes in the class path that the active JVM uses, and
3. OpenGL which is required for final visualisation.

While not required, it is highly recommended to use Java's -Xmx96m option that allows for additional heap memory to be used by the virtual machine.

---

[3] This disk can be downloaded from http://users.ox.ac.uk/~sjoh1188/3yp/DISK.zip
[4] Project code documentation is available online at http://users.ox.ac.uk/~sjoh1188/3yp/docs/